\documentclass[sigconf]{acmart}

\usepackage{booktabs} 
\usepackage{lipsum,multicol}

\copyrightyear{2018}
\acmYear{2018}
\setcopyright{acmlicensed}
\acmConference[GECCO '18]{Genetic and Evolutionary Computation Conference}{July 15--19, 2018}{Kyoto, Japan}
\acmBooktitle{GECCO '18: Genetic and Evolutionary Computation Conference, July 15--19, 2018, Kyoto, Japan}
\acmPrice{15.00}
\acmDOI{10.1145/3205455.3205523}
\acmISBN{978-1-4503-5618-3/18/07}

\editor{Jennifer B. Sartor}
\editor{Theo D'Hondt}
\editor{Wolfgang De Meuter}

\begin{document}
\title{Evolving Event-driven Programs with SignalGP}

\author{Alexander Lalejini}
\affiliation{
    \institution{BEACON Center for the Study of Evolution in Action}
	\institution{Michigan State University, USA}
}
\email{lalejini@msu.edu}

\author{Charles Ofria}
\affiliation{
	\institution{BEACON Center for the Study of Evolution in Action}
	\institution{Michigan State University, USA}
}
\email{ofria@msu.edu}

\begin{abstract}
We present SignalGP, a new genetic programming (GP) technique designed to incorporate the event-driven programming paradigm into computational evolution's toolbox. Event-driven programming is a software design philosophy that simplifies the development of reactive programs by automatically triggering program modules (event-handlers) in response to external events, such as signals from the environment or messages from other programs. SignalGP incorporates these concepts by extending existing tag-based referencing techniques into an event-driven context. Both events and functions are labeled with evolvable tags; when an event occurs, the function with the closest matching tag is triggered. In this work, we apply SignalGP in the context of linear GP. We demonstrate the value of the event-driven paradigm using two distinct test problems (an environment coordination problem and a distributed leader election problem) by comparing SignalGP to variants that are otherwise identical, but must actively use sensors to process events or messages.  In each of these problems, rapid interaction with the environment or other agents is critical for maximizing fitness. We also discuss ways in which SignalGP can be generalized beyond our linear GP implementation. 
\end{abstract}

%



\maketitle

\newcommand{\code}{\texttt}

\section{Introduction}
Here, we introduce SignalGP, a new genetic programming (GP) technique designed to provide evolution direct access to the event-driven programming paradigm, allowing evolved programs to handle signals from the environment or from other agents in a more biologically inspired way than traditional GP approaches. 
In SignalGP, signals (\textit{e.g.} from the environment or from other agents) direct computation by triggering the execution of program modules (\textit{i.e.} functions). SignalGP augments the tag-based referencing techniques demonstrated by Spector \textit{et al.} \citep{Spector_Tag_2011,Spector_What_2011,Spector_Tag_2012} to specify which function is triggered by a signal, allowing the relationships between signals and functions to evolve over time. 
The SignalGP implementation presented here is demonstrated in the context of linear GP, wherein programs are represented as linear sequences of instructions; however, the ideas underpinning SignalGP are generalizable across a variety of genetic programming representations. 

Linear genetic programs generally follow an imperative programming paradigm where computation is driven procedurally. Execution often starts at the top of a program and proceeds in sequence, instruction-by-instruction, jumping or branching as dictated by executed instructions \citep{brameier2007linear,McDermott_Springer_2015}. In contrast to the imperative programming paradigm, program execution in event-driven computing is directed primarily by signals (\textit{i.e.} events), easing the design and development of programs that, much like biological organisms, must react on-the-fly to signals in the environment or from other agents. Is it possible to provide similarly useful abstractions to evolution in genetic programming? 

Different types of programs are more or less challenging to evolve depending on how they are represented and interpreted.  
By capturing the event-driven programming paradigm, SignalGP targets problem domains where agent-agent and agent-environment interactions are crucial, such as in robotics or distributed systems. 

In the following sections, we provide a broad overview of the event-driven paradigm, discussing it in the context of an existing event-driven software framework, cell signal transduction, and an evolutionary computation system for evolving robot controllers. 
Next, we discuss our implementation of SignalGP in detail. 
Then, we use SignalGP to demonstrate the value of capturing event-driven programming in GP with two test problems: an environment coordination problem and a distributed leader election problem. 
Finally, we conclude with planned extensions, including how SignalGP can be generalized beyond our linear GP implementation to other forms of GP. 

\section{The event-driven paradigm}
The event-driven programming paradigm is a software design philosophy where the central focus of development is the processing of events \citep{etzion2011event,Heemels_An_2012,Cassandras_2014}. Events often represent messages from other agents or processes, sensor readings, or user actions in the context of interactive software applications. Events are processed by callback functions (\textit{i.e.} event-handlers) where the appropriate event-handler is determined by an identifying characteristic associated with the event, often the event's name or type. In this way, events can act as remote function calls, allowing external signals to direct computation. 

Software development environments that support the event-driven paradigm often abstract away the logistics of monitoring for events and triggering event-handlers, simplifying the code that must be designed and implemented by the programmer and easing the development of reactive programs.
Thus, the event-driven paradigm is especially useful when developing software where computation is most appropriately directed by external stimuli, which is often the case in domains such as robotics, embedded systems, distributed systems, and web applications. 

For any event-driven system, we can address the following three questions: What are events? How are event-handlers represented? And, how does the system determine the most appropriate event-handler to trigger in response to an event? 
Crosbie and Spafford \citep{crosbie1996evolving} have addressed why answering such questions can be challenging in genetic programming; thus, it is useful to look to how existing event-driven systems address them.
While many systems that exhibit event-driven characteristics exist, we restrict our attention to three: the Robot Operating System (ROS) \citep{quigley2009ros}, the biological process of signal transduction, and Byers \textit{et al.}'s digital enzymes robot controller \citep{byers2011digital,Byers_Exploring_2012}. 

ROS is a popular robotics software development framework that provides standardized communication protocols to independently running programs referred to as nodes. While the ROS framework provides a variety of tools and other conveniences to robotics software developers, we focus on ROS's publish-subscribe communication protocol, framing it under the event-driven paradigm. ROS nodes can communicate by passing strictly typed messages over named channels (topics). Nodes send messages by publishing them over topics, and nodes receive messages from a particular topic by subscribing to that topic. A node subscribes to a topic by registering a callback function that takes the appropriate message type as an argument. Anytime a message is sent over a topic, all callback functions registered with the topic are triggered, allowing subscribed nodes to react to published messages. Topics can have any number of publishers and subscribers, all agnostic to one another \citep{quigley2009ros}. In ROS's publish-subscribe system, events are represented as strictly typed messages, event-handlers are callback functions that take event information as input, and named channels (topics) determine which event-handlers an event triggers. 
 
The behavior of many natural systems can be interpreted as using the event-driven paradigm. 
In cell biology, signal transduction is the process by which a cell transforms an extracellular signal into a response, often in the form of cascading biochemical reactions that alter the cell's behavior. Cells respond to signaling molecules via receptors, which bind specifically to nearby signaling molecules and initiate the cell's response \citep{alberts2017molecular}. The process of cell signal transduction can be viewed as a form of event-driven computation: signaling molecules are like events, receptors are event-handlers, and the chemical and physical properties of signaling molecules determine with which receptors they are able to bind. 

Evolutionary computation researchers have also made use of the event-driven paradigm; for example, Byers \textit{et al.} \citep{byers2011digital,Byers_Exploring_2012} demonstrated virtual robot controllers that operate using a digital model of signal transduction, and like biological signal transduction, these controllers follow an event-driven paradigm. Byers \textit{et al.}'s virtual robot controllers have digital stimuli receptors, which bind to nearby ``signaling molecules'' represented as bit strings. Different bit strings represent different signals in the environment (\textit{e.g.} the presence of nearby obstacles). Once a signaling molecule binds to a digital receptor, a digital enzyme (program) processes the signaling molecule and influences the controller's behavior. In a single controller, there are many digital enzymes (not all of the same type) processing signaling molecules in parallel, all vying to influence the controller's actions; in this way, virtual robot behavior emerges. As in cell signal transduction, signaling molecules are events, digital stimuli receptors and digital enzymes act as event-handlers, and events are paired with handlers based on signal type and signal location.

\section{SignalGP}
\begin{figure*}[!ht]
   \includegraphics[width=\textwidth]{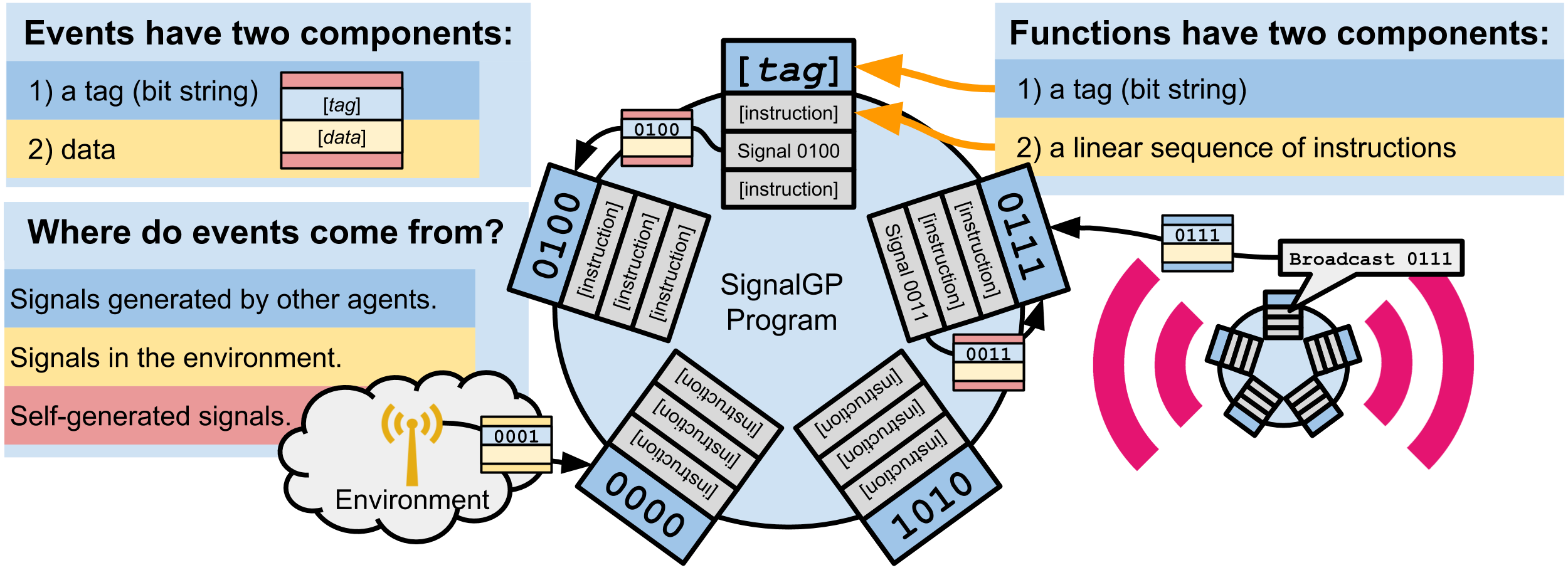}
  \caption{\small A high-level overview of SignalGP. SignalGP programs are defined by a set of functions. Events trigger functions with the \textit{closest matching} tag, allowing SignalGP agents to respond to signals. SignalGP agents handle many events simultaneously by processing them in parallel. }
  \label{fig:sgp_overview_cartoon}
\end{figure*}

As with other tag-based systems, 
SignalGP agents (programs) are defined by a set of functions (modules) where each function is referred to using a tag and contains a linear sequence of instructions.
To augment this framework, SignalGP also makes explicit the concept of \textit{events} where event-specific data is associated with a tag that agents can use to specify how that event should be handled.  
In this work, we arbitrarily chose to represent tags as fixed-length bit strings. 
Agents may both generate internal events and be subjected to events generated by the environment or by other agents. 
Events trigger functions based on the similarity of their tags. When an event triggers a function, the function is run with the event's associated data as input. 
SignalGP agents handle many events simultaneously by processing them in parallel. 
Figure \ref{fig:sgp_overview_cartoon} shows a high-level overview of SignalGP. 

\subsection{Tag-based Referencing} 
Incorporating modules (\textit{e.g.} functions, subroutines, macros, \textit{etc.}) into genetic programming has been extensively explored, and the benefits of modules in GP have been well documented (\textit{e.g.} \citep{koza1994genetic,koza1992genetic,angeline1992evolutionary,keijzer2005undirected,Walker_The_2008,roberts2001evolving,spector1996simultaneous}). The main purpose of SignalGP functions are to act as event-handlers --- computations triggered in response to signals. However, they have the additional benefit of providing explicit architectural support for program modularity, bestowing the boon of reusable code. 
As with any reusable code block in GP, the question remains: how should the code be referenced? The answer to this question can be reused to answer the following question: how should we determine which event-handlers are triggered by events? 

Inspired by John Holland's concept of a ``tag'' (\citep{holland1993effect,holland1987genetic,holland1990concerning,holland2006studying}) as a mechanism for matching, binding, and aggregation, Spector \textit{et al.} \citep{Spector_Tag_2011,Spector_What_2011,Spector_Tag_2012} introduced and demonstrated the value of tag-based referencing in the context of GP. 
In this context, a tag-based reference always links to a tagged entity with its closest match. 
These tagged entities include instructions and sequences of code (\textit{i.e.} modules), providing an evolvable mechanism for code referencing. 

SignalGP shifts these ideas into a more fully event-driven context.
In SignalGP, sets of instructions are modularized into functions that are labeled with tags. Events are made explicit and trigger those functions with whose tags have the closest match. The underlying instruction set is crafted to easily trigger internal events, broadcast external events, and to otherwise work in a tag-based context. 
Finally, SignalGP can be configured to only match tags that are relatively close (within a threshold) allowing agents to ignore events entirely by avoiding the use of similar tags.

\subsection{Virtual Hardware} 
As in many GP representations, linear GP programs are often interpreted in the context of virtual hardware, which typically comprises memory --- usually in the form of registers or stacks --- and other problem-specific virtual hardware elements, allowing programs to achieve complex functionality \citep{McDermott_Springer_2015,McPhee_Field_2008,Ofria_Avida_2009}. 
SignalGP programs are interpreted by virtual hardware consisting of the following four major components: program memory, an event queue, 
a set of execution threads, and shared memory.

\textbf{Program memory} stores the SignalGP program currently executing on the virtual hardware.

The \textbf{event queue} manages recently received events waiting to be dispatched and processed by functions. The event queue dispatches events in the order they are received. 

The SignalGP virtual hardware supports an arbitrary number of \textbf{execution threads} that run concurrently. Each thread processes a single instruction every time step. In the same way that Byers \textit{et al.}'s parallel-executing digital enzymes \citep{byers2011digital} allow a robot controller to process many external stimuli simultaneously, parallel execution allows SignalGP agents to handle many events at once.

Each thread maintains a call stack that stores state information about the thread's active function calls. The current state for any given thread resides at the top of the thread's call stack. Call states maintain local state information for the function call they represent: a function pointer, an instruction pointer, input memory, working memory, and output memory. A function pointer indicates the current function being run. An instruction pointer indicates the current instruction within that function. Input, working, and output memory serve as local memory. 

Working memory is used for performing local operations (\textit{e.g.} addition, subtraction, multiplication, \textit{etc.}). Input memory is analogous to function arguments (\textit{i.e.} function input), and output memory is analogous to function return memory (\textit{i.e.} what is returned when a function call concludes). 
By convention, instructions can both read from and write to working memory, input memory is read-only, and output memory is write-only. 
To use an analogy, working memory, input memory, and output memory are to SignalGP functions as hidden nodes, input nodes, and output nodes are to conventional artificial neural networks.  

\textbf{Shared memory} serves as global memory. Shared memory is accessible (\textit{i.e.} readable and writable) by all threads, allowing them to store and share information. 

\subsection{Program Evaluation}
SignalGP programs are sets of functions where each function associates an evolvable tag with a linear sequence of instructions. 
In our implementation of SignalGP, instructions are argument-based, and in addition to evolvable arguments, each instruction has an evolvable tag. Arguments modify the effect of the instruction, often specifying memory locations or fixed values. Instruction tags may also modify the effect of an instruction. For example, instructions that refer to functions do so using tag-based referencing. 
Further, instructions use their tag when generating events, either to be broadcast to other SignalGP agents or to be handled internally for their own use. 

Program evaluation can be initialized either actively or passively.  During active initialization, the program will begin evaluation by automatically calling a designated main function on a new thread.  In passive initialization, computation takes place only in response to external events.  In the work presented here, we use only active initialization and automatically reset the main thread if it would have otherwise terminated.

While executing, the SignalGP virtual hardware advances on each time step in three phases: 
(1) All events in the event queue are dispatched, with each triggering a function \textit{via} tag-based referencing. 
(2) Each thread processes a single instruction. 
(3) Any threads done processing are removed.
Phases occur serially and in order. 

Executed instructions may call functions, manipulate local and shared memory, generate events, perform basic computations, control execution flow, \textit{et cetera} (see supplementary material for details on all instructions used in this work).
Instructions in SignalGP are guaranteed to always be syntactically valid, but may be functionally useless. 
Every instruction has three associated arguments and an associated tag. 
Not all instructions make use of their three arguments or their tag; unused arguments and tags are not under direct selection and may drift until a mutation to the operand reveals them.

\vspace{2mm}
\noindent\textbf{Instruction-triggered Function Calls}\\
Functions in SignalGP may be triggered by either instruction calls or events.
When a \code{Call} is executed, the function in program memory with the most similar tag to the \code{Call} instruction's tag (above a similarity threshold) is triggered; in this work, ties are broken by a random draw (though any tie-breaking procedure could be used).  
Tag similarity is calculated as the proportion of matching bits between two bit strings (simple matching coefficient).  

When a function is triggered by a \code{Call} instruction, a new call state is created and pushed onto that thread's call stack. The working memory of the caller state is copied as the input memory of the new call state (\textit{i.e.} the arguments to the called function are the full contents of the previous working memory). The working memory and the output memory of the new call state are initially empty. To prevent unbounded recursion, we place limits on call stack depth; if a function call would cause the call stack to exceed its depth limit, the call instead behaves like a no-operation. 

Instruction-triggered functions may return by either executing a \code{Return} instruction or by reaching the end of the function's instruction sequence. When an instruction-triggered function returns, its call state is popped from its call stack, and anything stored in the output memory of the returning call state is copied to the working memory of the caller state (otherwise leaving the caller state's working memory unchanged). In this way, instruction-triggered function calls can be thought of as operations over the caller's working memory. 

\vspace{2mm}
\noindent\textbf{Event-triggered Function Calls}\\
Events in SignalGP are analogous to external function calls. When an event is dispatched from the event queue, the virtual hardware chooses the function with the highest tag similarity score (above a similarity threshold) to handle the event; in this work, ties are broken by a random draw (though any tie-breaking procedure could be used). 

Once a function is selected to handle an event, it is called on a newly-created execution thread, initializing the thread's call stack with a new call state. The input memory of the new call state is populated with the event's data. In this way, events can pass information to the function that handles them. When the function has been processed (\textit{i.e.} all of the active calls on the thread's call stack have returned), the thread is removed. To prevent unbounded parallelism, we place a limit on the allowed number of concurrently executing threads; if the creation of a new thread would cause the number of threads to exceed this limit, thread creation is prevented.

\subsection{Evolution}
\label{sec:evolution}
Evolution in SignalGP proceeds similarly to that of typical linear GP systems. 
Because function referencing is done \textit{via} tags, changes can be made to program architecture (\textit{e.g.} inserting new or removing existing functions) while still guaranteeing syntactic correctness. 
Thus, modular program architectures can evolve dynamically through whole-function duplication and deletion operators or through function-level crossover techniques. 

In the studies presented in this paper, we evolve SignalGP programs directly (as opposed to using indirect program encodings), which requires SignalGP-aware mutation operators. 
We propagated SignalGP programs asexually and applied mutations to offspring. 
We used whole-function duplication and deletion operators (applied at a per-function rate of 0.05) to allow evolution to tune the number or functions in programs. 
We mutated tags for instructions and functions at a per-bit mutation rate (0.05). 
We applied instruction and argument substitutions at a per-instruction rate (0.005). 
Instruction sequences could be inserted or deleted via slip-mutation operators \citep{lalejini_gene_2017}, which facilitate the duplication or deletion of sequences of instructions; we applied slip-mutations at a per-function rate (0.05). 

SignalGP is under active development as part of the Empirical library ({\small \url{https://github.com/devosoft/Empirical}}).

\section{Test Problems}
We demonstrate the value of incorporating the event-driven programming paradigm in GP using two distinct test problems: a changing environment problem and a distributed leader-election problem. For both problems, we compared SignalGP performance to variants that are otherwise identical, except for how they handle sensor information. 
For example, our primary variant GP must actively monitor sensors to process external signals (using the imperative paradigm).
For both test problems, a program's capacity to react efficiently to external events is crucial; thus, we hypothesized that SignalGP should perform better than our imperative alternatives.  

\subsection{Changing Environment Problem}
This first problem requires agents to coordinate their behavior with a randomly changing environment. The environment can be in one of $K$ possible states; to maximize fitness, agents must match their internal state to the current state of their environment. The environment is initialized to a random state and has a 12.5\% chance of changing to a random state at every subsequent time step.
Successful agents must adjust their internal state whenever an environmental change occurs. 

We evolved agents to solve this problem at $K$ equal to 2, 4, 8, and 16 environmental states. The problem scales in difficulty as the number of possible states that must be monitored increases. Agents adjust their internal state by executing one of $K$ state-altering instructions.
For each possible environmental state, there is an associated \code{SetState} instruction
(\textit{i.e.} for $K=4$, there are four instructions: \code{SetState0}, \code{SetState1}, \code{SetState2}, and \code{SetState3}).
Being required to execute a distinct instruction for each environment represents performing a behavior unique to that environment. 

We compared the performance of programs with three different mechanisms to sense the environment: (1) an event-driven treatment where environmental changes produce signals that have environment-specific tags and can trigger functions; (2) an imperative control treatment where programs needed to actively poll the environment to determine its current state; and (3) a combined treatment where agents are capable of using either option.  Note that in the imperative and combined treatments we added new instructions to test each environmental state (\textit{i.e.} for $K=4$, there are four instructions: \code{SenseEnvState0}, \code{SenseEnvState1}, \code{SenseEnvState2}, and \code{SenseEnvState3}). 
In preliminary experiments we had provided agents with a single instruction that returned the current environmental state, but this mechanism proved more challenging for them to use effectively when there were too many states (the environment ID returned by the single instruction needed to be thresholded into a true/false value, whereas the individual environment state tests directly returned a true/false value). 
Across all treatments, we added a \code{Fork} instruction to the available instruction set. The \code{Fork} instruction generates an internally-handled signal when executed, which provides an independent mechanism to spawn parallel-executing threads. The \code{Fork} instruction ensures that programs in all treatments had trivial access to parallelism. Because the \code{SenseEnvState} instructions in both the imperative and combined treatments bloated the instruction set relative to the event-driven treatment, we also added an equivalent number of no-operation instructions in the event-driven treatment.

\subsubsection{Hypotheses}
For low values of $K$, we expected evolved programs from all treatments to perform similarly. However, as continuously polling the environment is  cumbersome at higher values of $K$, we expected fully event-driven SignalGP programs to drastically outperform programs evolved in the imperative treatment; further, we expected successful programs in the combined treatment to favor the event-driven strategy. 

\subsubsection{Experimental Parameters}
We ran 100 replicates of each condition at $K$ = 2, 4, 8, and 16. In all replicates and across all treatments, we evolved populations of 1000 agents for 10,000 generations, starting from a simple ancestor program consisting of a single function with eight no-operation instructions. 
Each generation, we evaluated all agents in the population individually three times (three trials) where each trial comprised 256 time steps. 
For a single trial, an agent's fitness was equal to the number of time steps in which its internal state matched the environment state during evaluation. 
After three trials, an agent's fitness was equal to the minimum fitness value obtained across its three trials. 
We used a combination of elite and tournament (size four) selection to select which individuals reproduced asexually each generation. We applied mutations to offspring as described in Section \ref{sec:evolution}. 
Agents were limited to a maximum of 32 parallel executing threads and a maximum of 32 functions. 
Functions were limited to a maximum length of 128 instructions. 
Agents were limited to 128 call states per call stack. The minimum tag reference threshold was 50\% (\textit{i.e.} tags must have at least 50\% similarity to successfully reference). All tags were represented as length 16 bit strings. 

\subsubsection{Statistical Methods}
For every run, we extracted the program with the highest fitness after 10,000 generations of evolution. Because the sequence of environmental states experienced by an agent during evaluation are highly variant, we tested each extracted program in 100 trials, using a program's average performance as its fitness in our analyses. For each environment size, we compared the performances of evolved programs across treatments. To determine if any of the treatments were significant ($p < 0.05$) within a set, we performed a Kruskal-Wallis test. For an environment size in which the Kruskal-Wallis test was significant, we performed a post-hoc Dunn's test, applying a Bonferroni correction for multiple comparisons. All statistical analyses were conducted in R 3.3.2 \citep{r_core_2016}, and each Dunn's test was done using the FSA package \citep{r_FSA_2017}.

\subsection{Distributed Leader Election Problem}
In the distributed leader election problem, a network of agents must unanimously designate a single agent as leader. Agents are each given a unique identifier (UID). Initially, agents are only aware of their own UID and must communicate to resolve the UIDs of other agents. During an election, each agent may vote, and an election is successful if all votes converge to a single, consensus UID. 
This problem has been used to study the evolution of cooperation in digital systems \citep{Knoester_Using_2007,knoester2013genetic} and as a benchmark problem to compare the performance of different GP representations in evolving distributed algorithms \citep{Weise_Evolving_2012}. A common strategy for successfully electing a leader begins with all agents voting for themselves. Then, agents continuously broadcast their vote, changing it only when they receive a message containing a UID greater than their current vote. This process results in the largest UID propagating through the distributed system as the consensus leader. Alternatively, a similar strategy works for electing the agent with the smallest UID. 

We evolved populations of homogeneous distributed systems of SignalGP agents where networks were configured as 5x5 toroidal grids, and agents could only interact with their four neighbors. When evaluating a network, we initialized each agent in the network with a random UID (a number between 1 and 1,000,000). We evaluated distributed systems for 256 time steps. 
During an evaluation, agents retrieve their UID by executing a \code{GetUID} instruction. Agents vote with a \code{SetOpinion} instruction, which sets their opinion (vote) to a value stored in memory, and agents may retrieve their current vote by executing a \code{GetOpinion} instruction. Agents communicate by exchanging messages, either by sending a message to a single neighbor or by broadcasting a message to all neighboring agents. 

After an evaluation, we assigned fitness, $F$ according to Equation \ref{equ:consensus_fitness} where 
$V$ gives the number of valid votes at the end of evaluation, 
$C_{\text{max}}$ gives the maximum consensus size at the end of evaluation,  
$T_{\text{consensus}}$ gives the total number of time steps at full consensus, and
$S$ gives the size of the distributed system.

\begin{equation}
F = V + C_{\text{max}} + (T_{\text{consensus}} \times S)
\label{equ:consensus_fitness}
\end{equation}

Distributed systems maximize their fitness by achieving consensus as quickly as possible and maintaining consensus for the duration of their evaluation. Our fitness function rewards partial solutions by taking into account valid votes (\textit{i.e.} votes that correspond to a UID present in the network) and partial consensus at the end of an evaluation. 

We evolved distributed systems in three treatments: one with event-driven messaging and two different imperative messaging treatments. 
In the event-driven treatment, messages were events that, when received, could trigger a function. 
In both imperative treatments, messages did not automatically trigger functions; instead, messages were sent to an inbox and needed to be retrieved \textit{via} a \code{RetrieveMessage} instruction. The difference between the two imperative treatments was in how messages were handled once retrieved. 
In the fork-on-retrieve imperative treatment, messages act like an internally-generated event when retrieved from an inbox, triggering the function with the closest (above a threshold) matching tag on a new thread. 
In the copy-on-retrieve imperative treatment, messages are not treated as internal events when retrieved; instead, message contents are loaded into the input memory of the thread that retrieved the message. In the copy-on-retrieve imperative treatment, we augmented the available instruction set with the \code{Fork} instruction, allowing programs to trivially spawn parallel-executing threads. 

\subsubsection{Hypothesis}
Event-driven SignalGP agents do not need to continuously poll a message inbox to receive messages from neighboring agents, allowing event-driven programs to more efficiently coordinate. Thus, we expected distributed systems evolved in the event-driven treatment to outperform those evolved in the two imperative treatments. 

\subsubsection{Experimental Parameters}
We ran 100 replicates of each treatment. In all replicates of all treatments, we evolved populations of 400 homogeneous distributed systems for 50,000 generations. We initialized populations with a simple ancestor program consisting of a single function with eight no-operation instructions. Selection and reproduction were identical to that of the changing environment problem. Agents were limited to a maximum of 8 parallel executing threads. 
Agents were limited to a maximum of 4 functions, and function length was limited to a maximum 32 instructions. 
Agents were limited to 128 call states per call stack. The minimum tag reference threshold was 50\%. All tags were represented as length 16 bit strings. The maximum inbox capacity was 8. If a message was received and the inbox was full, the oldest message in the inbox was deleted to make room for the new message. 

\subsubsection{Statistical Methods}
For every replicate across all treatments, we extracted the program that produces the most fit distributed system after 50,000 generations of evolution. As in the changing environment problem, we compared treatments using a Kruskal-Wallis test, and if significant ($p < 0.05$), we performed a post-hoc Dunn's test, applying a Bonferroni correction for multiple comparisons.  


\section{Results and Discussion}

\subsection{Changing Environment Problem}

\begin{figure*}[!ht]
  \centering 
  \includegraphics[width=\textwidth]{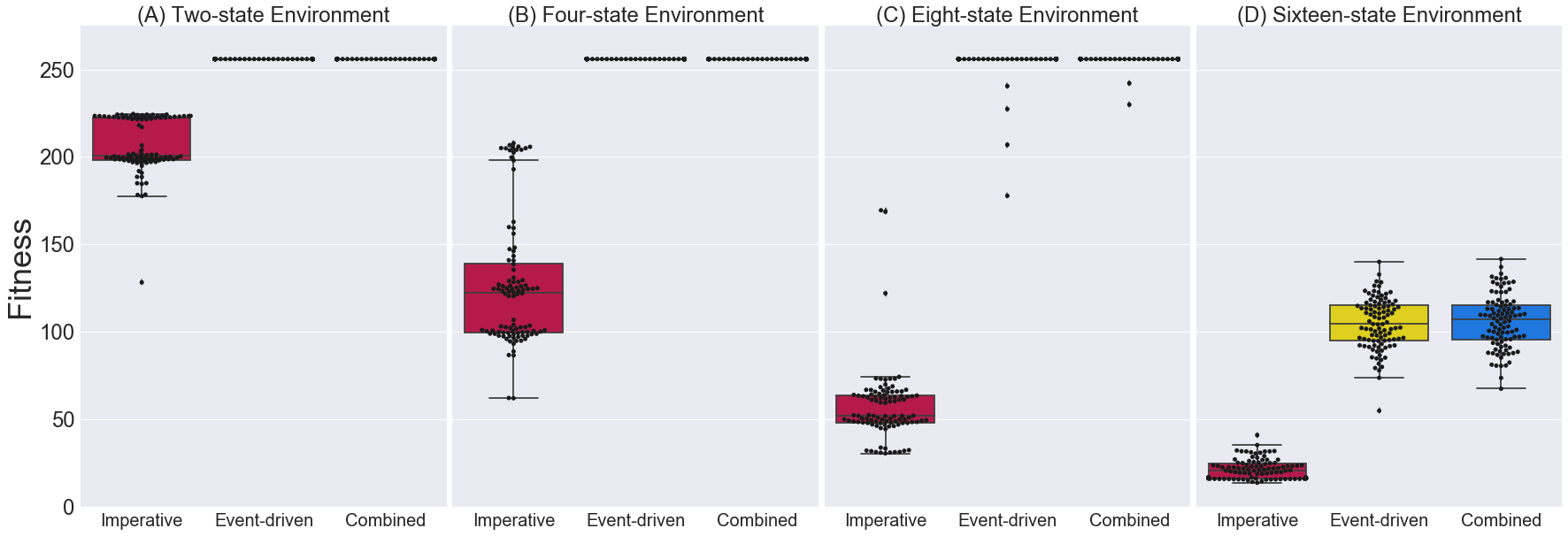}
  \caption{\small Changing environment problem results across all environments: A) two-state environment, B) four-state environment, C) eight-state environment, and D) sixteen-state environment. The box plots indicate the fitnesses (each an average over 100 trials) of best performing programs from each replicate.}
  \label{fig:chg_env_fitness}
\end{figure*}

\vspace{2mm}
\noindent\textbf{Event-driven strategies outperform imperative strategies.}\\
Figure \ref{fig:chg_env_fitness} shows results for all environment sizes ($K$ = 2, 4, 8, and 16). 
Programs evolved in treatments with fully event-driven SignalGP significantly outperformed those evolved in the imperative treatment across all environments: two-state (combined: $p=1.204471$e-47; event-driven: $p=1.204471$e-47), four-state (combined: $p=1.204512$e-47; event-driven: $p=1.204512$e-47), eight-state (combined: $p=1.283914$e-46; event-driven: $p=2.170773$e-45), and sixteen-state (combined: $p=1.906603$e-34; event-driven: $p=2.318351$e-33). Across all environments, there was no significant difference in final program performance between the event-driven and combined treatment. See supplementary material for full details on statistical test results. 

Further, only treatments with fully event-driven SignalGP produced programs capable of achieving a perfect fitness of 256. This result is not surprising: \textit{only} programs that employ an entirely event-driven strategy can achieve a perfect score in multi-state environments. This is because imperative strategies must continuously poll the environment for changes, which decreases the efficiency of their response to an environmental change. This strategy becomes increasingly cumbersome and inefficient as the complexity of the environment increases. 
In contrast, event-driven responses are triggered automatically \textit{via} the SignalGP virtual hardware, facilitating immediate reactions to environmental changes. This allows event-driven strategies to more effectively scale with environment size than imperative strategies. 

\vspace{2mm}
\noindent\textbf{Evolution favors event-driven strategies.}\\
In the combined treatment, evolution had access to both the event-driven (signal-based) strategy and the imperative (sensor-polling) strategy. As shown in Figure \ref{fig:chg_env_fitness}, performance in the combined treatment did not significantly differ from the event-driven treatment, but significantly exceeded performance in the imperative treatment. However, this result alone does not reveal what strategies were favored in the combined treatment. 

To tease this apart, we re-evaluated programs evolved under the combined treatment in two distinct conditions: one in which we deactivated sensors and one in which we deactivated external events. In the deactivated sensors condition, \code{SenseEnvState} instructions behaved as no-operations, which eliminated the viability of a sensor-based polling strategy. Likewise, the deactivated events re-evaluation condition eliminated the viability of event-driven strategies.  Any loss of functionality by programs in these new environments will tease apart the strategies that those programs must have employed.

\begin{figure}[h]
   \includegraphics[width=\columnwidth]{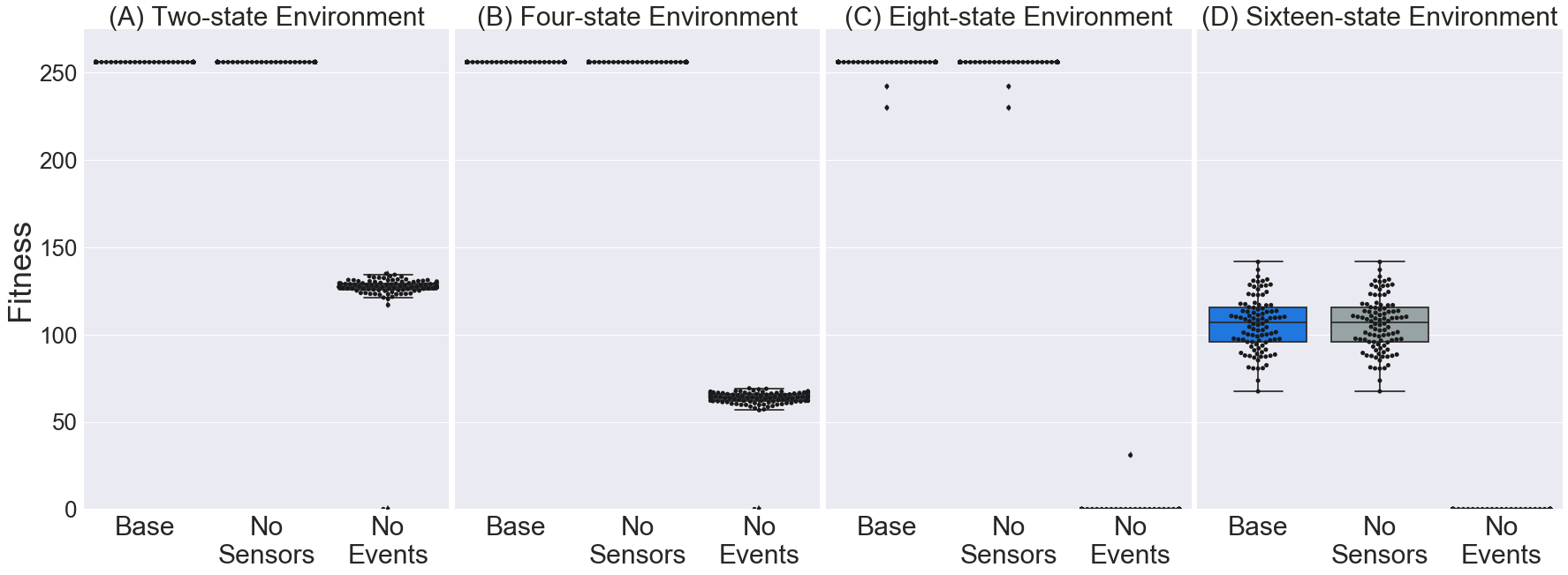}
   \caption{\small Re-evaluation results for combined condition in the changing environment problem across all environments: A) two-state environment, B) four-state environment, C) eight-state environment, and D) sixteen-state environment. The box plots indicate the fitnesses (each an average over 100 trials) of best performing programs from each re-evaluation.}
    \label{fig:combined_reval}
\end{figure}

Figure \ref{fig:combined_reval} shows the results of our re-evaluations. 
Across all environment sizes, there was no significant difference between program performance in their original combined condition and the deactivated sensors conditions. 
In contrast, program performances were significantly worse in the deactivated events condition than in the combined condition (two-state: $p=1.204228$e-47; four-state: $p=1.204106$e-47; eight-state: $p=6.743368$e-49 ; sixteen-state: $p=3.695207$e-35). 
These data indicate that programs evolved in the combined condition primarily rely on event-driven strategies for the changing environment problem.

\subsection{Distributed Leader Election Problem}

\begin{figure}[h]
   \includegraphics[width=\columnwidth]{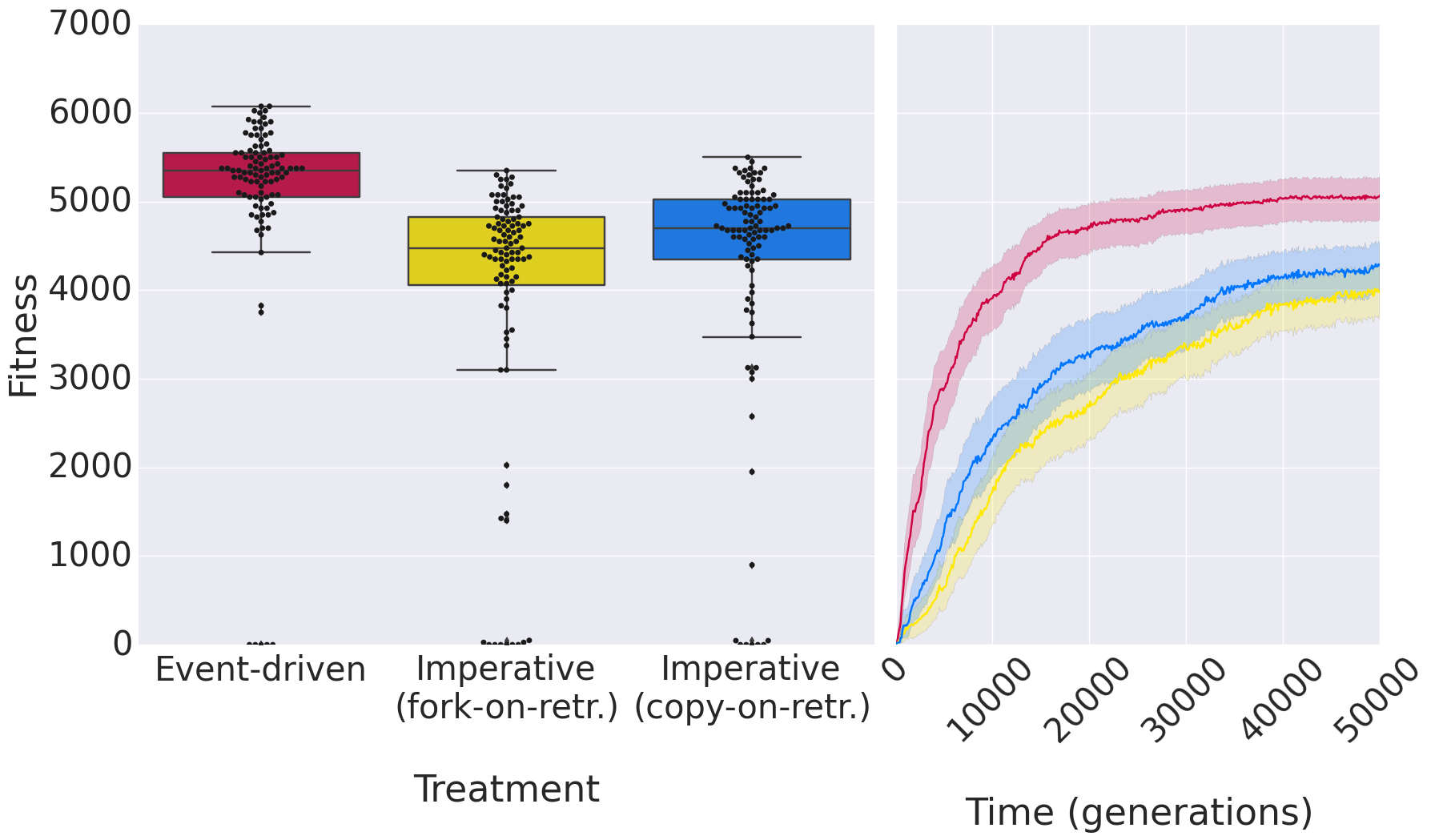}
  \caption{\small Distributed leader election problem results. The box plots indicate the fitnesses of best performing distributed systems from each replicate. The time series gives average fitness over time during evolution. The colors in the time series correspond to the colors in the box plots. The shading on fitness trajectories in the time series indicates a bootstrapped 95\% confidence interval.}
  \label{fig:consensus_fitness}
\end{figure}

\vspace{2mm}
\noindent\textbf{Event-driven networks outperform imperative networks.}\\
Figure \ref{fig:consensus_fitness} shows the results for the distributed leader election problem. Distributed systems evolved in the event-driven treatment significantly outperformed those evolved in both imperative treatments (fork-on-retrieval: $p=1.083074$e-21; copy-on-retrieval: $p=1.741302$e-13). See supplementary material for full details on statistical test results. 

All three conditions produced distributed systems capable of achieving election consensus. The difference in performances across treatments primarily reflect how quickly consensus is able to be reached within a distributed system. 
The event-driven programming paradigm is able to more efficiently encode communication between agents, as it does not require programs to continuously poll for new messages from other agents. 
Thus, the event-driven paradigm allows signals to propagate more quickly through a distributed system than the imperative paradigm. 
The time series shown in Figure \ref{fig:consensus_fitness} hints that the event-driven SignalGP representation evolves more rapidly for the distributed leader election problem than the imperative variants; however, deeper analyses are required for confirmation.

\section{Conclusion}
We introduced SignalGP, a new type of GP technique designed to provide evolution direct access to the event-driven programming paradigm by augmenting Spector \textit{et al.}'s \citep{Spector_Tag_2011} tag-based modular program framework. We have described and demonstrated SignalGP within the context of linear GP. Additionally, we used SignalGP to explore the value of capturing the event-driven paradigm on two problems where the capacity to react to external signals is critical: the changing environment problem, and the distributed leader election problem. At a minimum, our results show that access to the event-driven programming paradigm allows programs to more efficiently encode agent-agent and agent-environment interactions, resulting in higher performance on both the changing environment and distributed leader election problems. Deeper analyses are needed to tease apart the effects of the event-driven programming paradigm on the evolvability of solutions. 

\subsection{Beyond Linear GP}


While this work presents SignalGP in the context of linear GP, the ideas underpinning SignalGP are generalizable across a variety of evolutionary computation systems. 

We can imagine SignalGP functions to be black-box input-output machines. Here, we have exclusively put linear sequences of instructions inside these black-boxes, but could have easily put other representations capable of processing inputs (\textit{e.g.} other forms of GP, Markov brains \citep{hintze2017markov}, artificial neural networks, \textit{etc.}).
We could even employ black-boxes with a variety of different contents within the same agent. Encasing a variety of representations within a single agent may complicate the virtual hardware, program evaluation, and mutation operators, but also provides evolution with a toolbox of diverse representations.

As we continue to explore the capabilities of SignalGP, we plan to explore the evolvability of event-driven programs versus imperative programs across a wider set of problems and incorporate comparisons to other GP representations.  Further, we plan to extend SignalGP to other representations beyond linear GP and compare their relative capabilities and interactions.



\begin{acks}
We extend our thanks to Wolfgang Banzhaf and the members of the Digital Evolution Laboratory at Michigan State University for thoughtful discussions and feedback. 
This research was supported by the National Science Foundation (NSF) through the BEACON Center (Cooperative Agreement DBI-0939454), a Graduate Research Fellowship to AL (Grant No. DGE-1424871), and NSF Grant No. DEB-1655715 to CO.  Michigan State University provided computational resources through the Institute for Cyber-Enabled Research.
Any opinion, findings, and conclusions or recommendations expressed in this material are those of the authors and do not necessarily reflect the views of the NSF or MSU.
\end{acks}

\bibliographystyle{ACM-Reference-Format}
\bibliography{refs} 

\end{document}